\documentclass{article}
\usepackage{spconf,amsmath,graphicx}
\usepackage{tabularx}
\usepackage{amsfonts}

\title{DAM-GAN : Image Inpainting using Dynamic Attention Map based on Fake Texture Detection}
\name{Dongmin Cha \qquad Daijin Kim}
\address{Department of Computer Science and Engineering \\ Pohang University of Science and Technology, South Korea \\}

\newcolumntype{P}[1]{>{\centering\arraybackslash}p{#1}}

\begin{document}
\maketitle
\begin{abstract}
Deep neural advancements have recently brought remarkable image synthesis performance to the field of image inpainting. The adaptation of generative adversarial networks (GAN) in particular has accelerated significant progress in high-quality image reconstruction. However, although many notable GAN-based networks have been proposed for image inpainting, still pixel artifacts or color inconsistency occur in synthesized images during the generation process, which are usually called fake textures. To reduce pixel inconsistency disorder resulted from fake textures, we introduce a GAN-based model using dynamic attention map (DAM-GAN). Our proposed DAM-GAN concentrates on detecting fake texture and products dynamic attention maps to diminish pixel inconsistency from the feature maps in the generator. Evaluation results on CelebA-HQ and Places2 datasets with other image inpainting approaches show the superiority of our network.
\end{abstract}

\section{Introduction}
\label{sec:intro}
Image inpainting, or image completion, is a task about image synthesis technique aims to filling occluded regions or missing pixels with appropriate semantic contents. The main objective of image inpainting is producing visually authentic images with less semantic inconsistency using computer vision-based approaches. Traditional methods relied on a patch-based matching approach using the measurement of cosine similarity \cite{barnes2009patchmatch}. Recently, the remarkable capability of generative adversarial networks (GAN) \cite{goodfellow2014generative} has boosted image inpainting performance based on convolutional neural networks (CNN). Because of its hierarchical design, GAN with encoder-decoder structure has exceptional reconstruction ability compared to previous approaches. The decoder synthesizes visual images from the feature level as the encoder learns how to extract feature representations from images. Currently, GAN-based approaches constitute a dominant stream in image inpainting \cite{pathak2016context,iizuka2017globally,song2018contextual,yu2018generative,yu2019free,wang2020multistage}.

However, despite GAN's high image restoration performance, some pixel artifacts or color inconsistency called 'fake texture' inevitably occur in the process of decoding \cite{huang2020fakelocator,zhang2019detecting}. Fake pixels cause degradation of image restoration performance by dropping the appearance consistency in the synthesized image. To tackle this issue, we introduce dynamic attention map (DAM) that detects fake textures in feature map and highlights them by generating an attention mask (or attention map) \cite{zhang2019self} for image inpainting. Unlike existing GAN-based inpainting methods requiring high computational cost for generating attention map \cite{xie2019image,wang2020multistage}, our proposed DAM blocks exploit learnable convolutional layers for detecting fake texture and converting it into an attention map for each different scale of each decoding layer. We reported the comparisons on CelebA-HQ and Places2 datasets and showed that outcome of our DAM-GAN demonstrating higher quality than other existing inpainting methods including GAN-based approaches.

\begin{figure}
  \includegraphics[width=\linewidth]{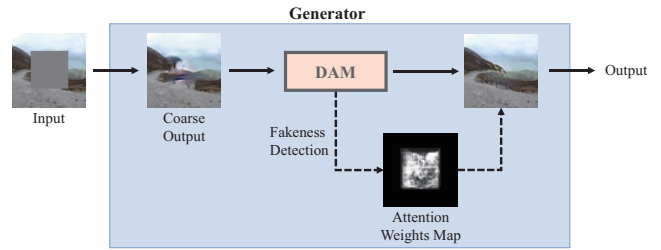}
  \caption{Overview of GAN-based image inpainting using our proposed dynamic attention map (DAM) module.}
  \label{fig:overview}
\end{figure}
\vspace{-0.1cm}
\begin{figure*}[t]
	\begin{center}
		\rule{0pt}{1ex}\hspace{1.24mm}\includegraphics[width=.85\linewidth]{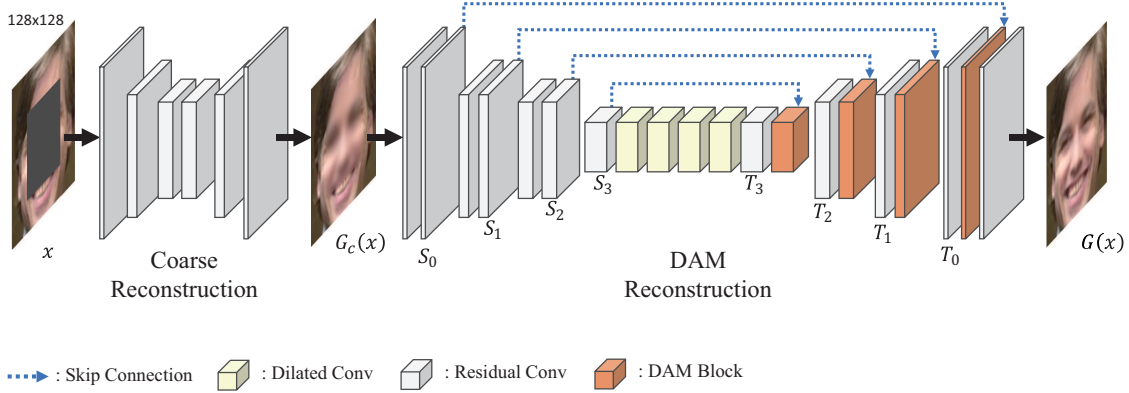}\\[-0.1pt]
	\end{center}
	\vspace{-0.2cm}
	\caption{Overall structure of generator $G$ in DAM-GAN. The coarse reconstruction part has a 3-level structure, and the DAM reconstruction part has a 4-level structure, including skip-connection and dynamic attention map (DAM) blocks.}
	\label{fig:overall_archi}
\end{figure*}

\section{Related Works}
\label{sec:format}
Traditional image inpainting methods were based on the exemplar-search approach, which divides image into patches to refill missing areas with other patches according to similarity computations such as PatchMatch \cite{barnes2009patchmatch}. Recently, progressive improvement of deep learning based generative models have demonstrated high feasibility for image synthesis. Especially GAN \cite{goodfellow2014generative} demonstrates brilliant performance in image inpainting tasks. Context Encoders (CE) \cite{pathak2016context} adopted encoder-decoder based GAN for image inpainting and Globally and Locally (GL) \cite{iizuka2017globally} incorporates global and local generators to maintain pixel consistency of output images. Contextual Attention (CA) \cite{yu2018generative} imitated the traditional patch-based method using GAN to take advantage of the basic concept of conventional exemplar-based methods. However, CE \cite{pathak2016context}, GL \cite{iizuka2017globally} and CA \cite{yu2018generative} have limitations on refilling irregular regions because of their local region based discriminators. Since they are usually specialized in reconstructing rectangular masks, images with free-shaped masks will decrease the quality of outputs. To tackle this limitations, recent inpainting approaches tend to remove local discriminator on architecture \cite{fang2020face}.

Partial conv \cite{liu2018image} did not employ GAN for inpainting, but solved the problem of generalization on irregular masks. It propose rule-based binary mask which is updated layer by layer in encoder-decoder network and showed high feasibility of refilling irregular masks. This mask-based inpainting approach is advanced in Gated conv \cite{yu2019free} by adopting GAN and replacing rule-based mask with learnable mask. Both Partial conv \cite{liu2018image} and Gated conv \cite{yu2019free} put forward a mask-based weights map for feature maps in the decoding process, similar to attention map \cite{zhang2019self} based method.

\section{Proposed Method}
Given original ground-truth $\hat{x}$, we apply occlusion mask to make input image $x$. The GAN-based inpainting network contains generator $G$ and discriminator $D$. Through the inpainting process in encoder-decoder architecture of generator, the output image $G(x)$ is obtained. In this section, we introduce our inpainting network with our proposed DAM module and loss functions for training our model.

\begin{figure*}[t]
	\begin{center}
		\rule{0pt}{1ex}\hspace{2.24mm}\includegraphics[width=.74\linewidth]{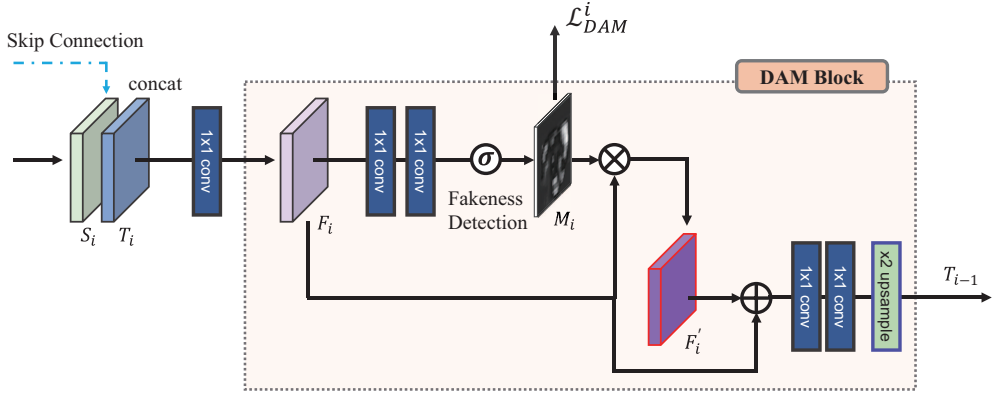}\\[-0.1pt]
	\end{center}
		\caption{An architecture of DAM block. Four DAM blocks are located in each layer of the decoder. Inside the i-th block, weight map ${M}_{i}$ is trained on ${L}_{DAM}$. $\otimes$ and $\oplus$ denote element-wise multiplication and summation between feature maps.}
	\label{fig:dam_module}
\end{figure*}

\subsection{GAN Framework}
The goal of generator $G$ is to fill missing parts with appropriate contents by understanding the input image $x$ (encoding) and synthesizing the output image $G(x)$ (decoding). Fig. \ref{fig:overall_archi} describes the overall architecture of generator $G$. The coarse reconstruction stage begins by filling pixels with a rough texture. The DAM reconstruction then uses DAM blocks to restore the coarse output ${G}_{C}(x)$ with detailed contents. We defined the residual convolution layer by combining residual block \cite{he2016deep} and convolution layer, and we adopted concatenation-based skip-connection \cite{ronneberger2015u} and dilated convolution \cite{chen2017deeplab} in the middle of the generator. Skip-connections have a notable effect on reducing vanishing gradient problems and maintaining spatial information of reconstructed images, and dilated convolution increases the receptive field to enhance the efficiency of the computations. 

Discriminator $D$ serves as a criticizer that distinguishes between real and synthesized images. Adversarial training between $G$ and $D$ can further improve the quality of synthesized image. Because local discriminator has critical limitations on handling irregular mask as mentioned in section 2., we use one global discriminator for adversarial training our model. We employed the global discriminator from CA \cite{yu2018generative}.

\subsection{Dynamic Attention Map Block}

Dynamic attention map (DAM) block located in each last four decoding layers from the generator $G$. The structure and function of DAM block are illustrated in Fig. \ref{fig:dam_module}. The concatenated feature $[{T}_{i}, {S}_{i}]$ passes through a 1x1 convolutional filter, and input feature ${F}_{i}$ is obtained.

\noindent
Similar to fakeness prediction in \cite{huang2020fakelocator}, fakeness map ${M}_{i}$ is produced through 1x1 convolutional filters and sigmoid function from feature ${F}_{i}$. Then, we can use ${M}_{i}$ as an attention map like \cite{zhang2019self}. After element-wise multiplication of ${M}_{i} \otimes {F}_{i}$, the output feature ${F'}_{i}$ is obtained. Then element-wise sum ${F}_{i} \oplus {F'}_{i}$ becomes the final output ${T}_{i-1}$, which is upsampled and passed to the upper layer in the decoder. Fakeness map ${M}_{i}$ is trainable dynamically in each layer from decoder using DAM loss $\mathcal{L}_{DAM}$, which is expressed in section 3.

\subsection{Loss Functions}

{\bf Reconstruction Loss \:\:} Image inpainting models are trained in pixel space by reducing the pixel-wise distance between ground-truth and synthesized images. We train both the coarse part and DAM part in the reconstruction process as shown in Eq. \ref{eq:reconstruction_loss}.

\begin{equation}
	\label {eq:reconstruction_loss}
	\mathcal{L}_{re} = |\hat{x} - {G}_{C}(x)|_{1} + |\hat{x} - G(x)|_{1}
\end{equation}

\noindent
Where $x$ and $\hat{x}$ represent masked image and ground-truth, ${G}_{C}(x)$ and $G(x)$ denote coarse and final output. %

\noindent
{\bf Adversarial Loss \:\:} %
Generator and discriminator are trained in a competitive relationship in a mini-max game to perform their roles. Generator $G$ tries to reconstruct the input image $x$ into inpainted image $G(x)$ as similar to the real image $\hat{x}$. Otherwise, discriminator $D$ is trained to distinguish real image $\hat{x}$ from fake image $x$ as possible. The adversarial loss is denoted in Eq. \ref{eq:adv_loss}. $D$ aims to maximize the adversarial loss $\mathcal{L}_{adv}$ while $G$ tries to minimize it.

\begin{equation}
	\label{eq:adv_loss}
	\mathcal{L}_{adv} = {\mathbb E}_{\hat{x}}[\log D(\hat{x})]  + {\mathbb E}_{x}[\log (1-D(G(x)))]
\end{equation}

\begin{figure*}[ht]
	\begin{center}
		\rule{0pt}{1ex}\hspace{2.24mm}\includegraphics[width=.85\linewidth]{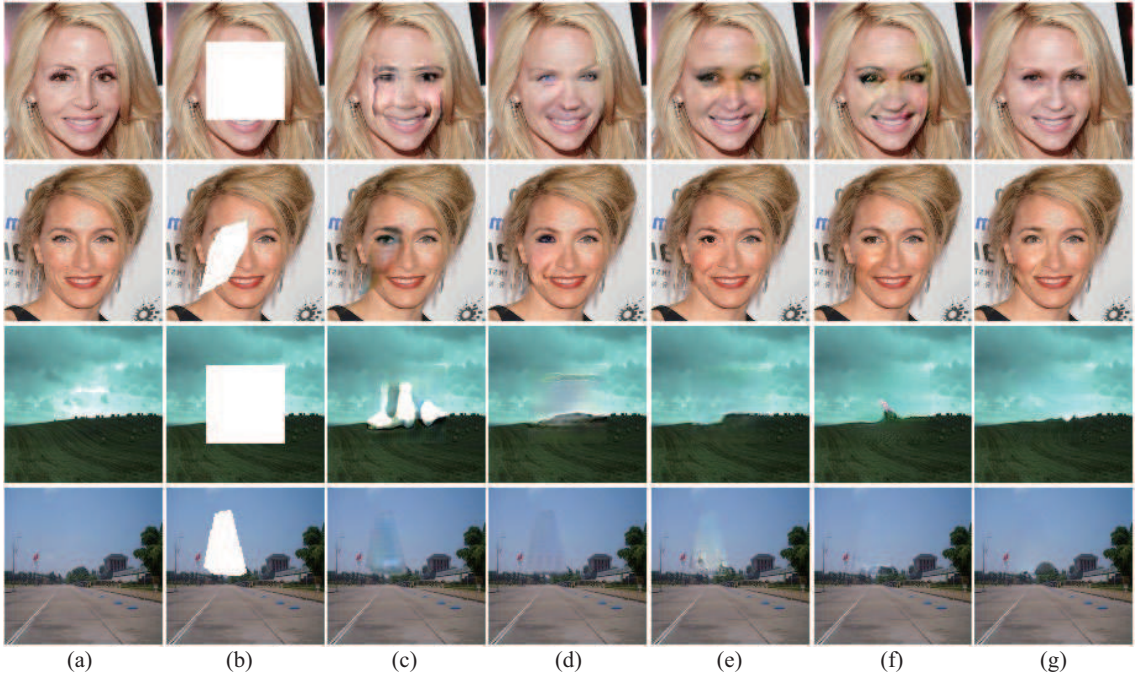}\\[-0.1pt]
	\end{center}
	\caption{Comparisons on CelebA-HQ \cite{karras2017progressive} and Places2 \cite{zhou2017places} using the centered mask and random masks. From left to right: (a) Ground-truth, (b) Input, (c) CE \cite{pathak2016context}, (d) CA \cite{yu2018generative}, (e) Partial \cite{liu2018image}, (f) Gated \cite{yu2019free} and (g) Our DAM-GAN.}
	\label{fig:results}
\end{figure*}

\noindent
{\bf DAM Loss \:\:} In each decoder layer, our proposed DAM block generates an attention map based on pixels from detected fake texture. We can consider those fake regions as pixels that will be highlighted during the image synthesis process. By computing the gray-scale pixel difference between real and inpainted images, the DAM block dynamically learns fake prediction functions from a ground-truth fakeness map. Then all pixels are divided by 255 to normalize them between [0, 1]. Formally, DAM loss can be described as in Eq. \ref{eq:dam_loss}.

\begin{equation}\label{eq:dam_loss}
  \begin{gathered}
    {M}^{GT}_{j} = grayscale (|{G(x)}_{j} - {\hat{x}}_{j}|) / 255.0        \\
    \mathcal{L}^{j}_{DAM} = |{M}_{j} - {M}^{GT}_{j}|_{1} \qquad
    \mathcal{L}_{DAM} = \sum_{j=0}^{3} \mathcal{L}^{j}_{DAM}
  \end{gathered}
\end{equation}

\noindent
Where $j \subseteq \{0, 1, 2, 3\}$, the j-th ground-truth mask ${M}^{j}_{GT}$ is obtained from the difference between real image (${\hat{x}}^{j}$) and inpainted image (${G(x)}^{j}$) resized with the same size of ${M}^{j}$.
\newline
\begin{table}[b!]
\vspace{0.4cm}
  \caption{Quantitative comparison results of image inpainting with CE  \cite{pathak2016context}, CA \cite{yu2018generative}, Partial \cite{liu2018image}, Gated \cite{yu2019free} and our DAM-GAN on CelebA-HQ dataset. The highest performances are marked in bold.}
  \vspace{0.2cm}
\begin{tabular}{l|P{1.35cm}P{1.35cm}P{1.35cm}P{1.35cm}}
\hline \hline

      & \multicolumn{2}{c}{Center}                          & \multicolumn{2}{c}{Free}                            \\ \cline{2-5} 
      & \multicolumn{1}{c}{\: PSNR} & \multicolumn{1}{c}{SSIM} & \multicolumn{1}{c}{PSNR} & \multicolumn{1}{c}{SSIM} \\ \hline 
CE & \: 22.56 & 0.864 & 27.20 & 0.939 \\
CA & \: 23.06 & 0.875 & 28.23 & 0.954 \\
Partial & \: 23.57 & 0.884 & 29.04 & 0.952 \\
Gated & \: 24.04 & 0.892 & 29.11 & 0.952 \\
{\bf Ours}  & \: {\bf 24.77} & {\bf 0.903} & {\bf 29.49} & {\bf 0.960} \\  \hline \hline
\end{tabular}
\label{tab:quantity}
\end{table}

\noindent
{\bf Full Objective Loss \:\:} The total loss function of DAM-GAN is defined as in Eq. \ref{eq:total_loss}. The hyper-parameters ${\lambda}_{re}, {\lambda}_{adv}$ and ${\lambda}_{DAM}$ denotes predefined weights for each component.

\begin{equation}
	\label {eq:total_loss}
	\mathcal{L}_{total} = {\lambda}_{re} \cdot \mathcal{L}_{re} + {\lambda}_{adv} \cdot \mathcal{L}_{adv} + {\lambda}_{DAM} \cdot \mathcal{L}_{DAM}
\end{equation}

\section{Experiments}

\subsection{Implementation Details}

Our model was trained on two datasets: CelebA-HQ and \cite{karras2017progressive} Places2 \cite{zhou2017places}. We randomly divided the 30,000 images in CelebA-HQ dataset into a training set of 27,000 images and a validation set of 3,000 images. In Places2 dataset, we select same categories as \cite{liu2021pd} in training set and tested our model on validation set. All images are resized to 128 $\times$ 128.
\noindent
To prepare input images for our model, we defined the centered mask and random mask. The centered mask has 64 $\times$ 64 size fixed in the center of the image, and the random mask has an irregular shape following the mask generation approach in \cite{park2021acn}. We used an ADAM optimizer \cite{kingma2014adam} in this experiment, and hyper-parameters are set to ${\lambda}_{re}=1, {\lambda}_{adv}=0.001$ and ${\lambda}_{DAM}=0.005$. %

\subsection{Quantitative Results}

As mentioned in CA \cite{yu2018generative}, image inpainting lacks a definable numerical metric for assessing inpainting results. Therefore, evaluation metrics in this field are still limited to pixel-level image quality metrics. The most commonly used quality comparison metrics in GAN tasks are the peak signal-to-noise ratio (PSNR) and the structural similarity index (SSIM) \cite{wang2004image}. We conducted quantitative comparisons on CelebA-HQ as shown in Table \ref{tab:quantity}. with four state-of-the-art inpainting benchmarks: {\bf CE} \cite{pathak2016context}, {\bf CA} \cite{yu2018generative}, {\bf Partial} \cite{liu2018image} and {\bf Gated} \cite{yu2019free}. However, since the local discriminator of {\bf CA} \cite{yu2018generative} cannot handle random masks, we conducted all experiments using only a global discriminator.

\subsection{Qualitative Results}

We displayed the results of our image inpainting and other four image inpainting approaches on CelebA-HQ and Places2 in Fig. \ref{fig:results}. It shows that our DAM-GAN trained with a dynamic attention map based on fakeness detection produces better visual image inpainting results than other models.

\section{CONCLUSION}
In this paper, we proposed a deep image inpainting generative model with dynamic attention map (DAM) blocks for weighting specific pixels in a feature map based on fake detection map. During training, the network learns itself how to refill missing regions using dynamically-learnable maps. We demonstrated that our DAM-GAN outperformed other inpainting models in terms of inpainting performance by comparing quantitative results.

\section{ACKNOWLEDGEMENTS}
\vspace{-0.2cm} 
This work was supported by Institute of Information \& communications Technology Planning \& Evaluation(IITP) grant funded by the Korea government(MSIT) (No.B0101-15-0266, Development of High Performance Visual BigData Discovery Platform for Large-Scale Realtime Data Analysis), (No.2017-0-00897, Development of Object Detection and Recognition for Intelligent Vehicles) and (No.2018-0-01290, Development of an Open Dataset and Cognitive Processing Technology for the Recognition of Features Derived From Unstructured Human Motions Used in Self-driving Cars)

\vfill\pagebreak

\bibliographystyle{IEEEbib}
\bibliography{damgan}

\end{document}